\documentclass[11pt,a4paper]{article}
\usepackage[hyperref]{acl2019}
\usepackage{times}
\usepackage{latexsym}
\usepackage{todonotes}

\usepackage{url}
\usepackage{booktabs}
\usepackage{amssymb}
\usepackage{amsmath}

\usepackage[percent]{overpic}
\usepackage{graphbox}

\usepackage{subcaption}
\usepackage{tikz}
\usetikzlibrary{positioning, arrows, shapes, calc}
\usepackage{multirow}
\usepackage{pgfplots}
\pgfplotsset{compat=1.14}

\tikzset{>=latex,
		 every pin edge/.style={shorten <=0.5mm}}
\usepgfplotslibrary{colorbrewer}
\pgfplotsset{
	cycle list/Set1-8,
	colormap/Set1-8,
	cycle multiindex* list={
		mark list*\nextlist
		Set1-8\nextlist
	},
	/pgfplots/bar cycle list/.style={/pgfplots/cycle list={
	{index of colormap=0, fill=.},
	{index of colormap=1, fill=.},
	{index of colormap=2, fill=.},
	{index of colormap=3, fill=.},
	{index of colormap=4, fill=.},
	{index of colormap=5, fill=.},
	{index of colormap=6, fill=.},
	{index of colormap=7, fill=.}
	}%
    },
}

\aclfinalcopy 
\def\aclpaperid{14} 

\allowdisplaybreaks

\title{Learning Bilingual Sentence Embeddings via Autoencoding\\and Computing Similarities with a Multilayer Perceptron}

\author{Yunsu Kim$^{2}$ \hspace{8pt} Hendrik Rosendahl$^{1,2}$ \hspace{8pt} Nick Rossenbach$^{2}$\\
  \textbf{Jan Rosendahl$^{2}$ \hspace{8pt} Shahram Khadivi$^{1}$ \hspace{8pt} Hermann Ney$^{2}$}\\
  $^{1}$eBay, Inc., Aachen, Germany\\
  {\tt \{hrosendahl,skhadivi\}@ebay.com}\\
  $^{2}$RWTH Aachen University, Aachen, Germany\\
  {\tt \{surname\}@cs.rwth-aachen.de}\\}

\date{}

\begin{document}
\maketitle
\begin{abstract}
  We propose a novel model architecture and training algorithm to learn bilingual sentence embeddings from a combination of parallel and monolingual data. Our method connects autoencoding and neural machine translation to force the source and target sentence embeddings to share the same space without the help of a pivot language or an additional transformation.
  We train a multilayer perceptron on top of the sentence embeddings to extract good bilingual sentence pairs from nonparallel or noisy parallel data. Our approach shows promising performance on sentence alignment recovery and the WMT 2018 parallel corpus filtering tasks with only a single model.\\
\end{abstract}

\section{Introduction}

Data crawling is increasingly important in machine translation (MT), especially for neural network models.
Without sufficient bilingual data, neural machine translation (NMT) fails to learn meaningful translation parameters \cite{koehn2017six}.
Even for high-resource language pairs, it is common to augment the training data with web-crawled bilingual sentences to improve the translation performance \cite{bojar-EtAl:2018:WMT1}.

Using crawled data in MT typically involves two core steps: mining and filtering.
Mining parallel sentences, i.e. aligning source and target sentences, is usually done with lots of heuristics and features: document/URL meta information \cite{resnik2003web,esplagomis2009bitextor}, sentence lengths with self-induced lexicon \cite{moore2002fast,varga2005parallel,etchegoyhen2016set}, word alignment statistics and linguistic tags \cite{stefuanescu2012hybrid,kaufmann2012jmaxalign}.

Filtering aligned sentence pairs also often involves heavy feature engineering \cite{taghipour2011parallel,xu2017zipporah}.
Most of the participants in the WMT 2018 parallel corpus filtering task use large-scale neural MT models and language models as the features \cite{koehn2018findings}.

Bilingual sentence embeddings can be an elegant and unified solution for parallel corpus mining and filtering.
They compress the information of each sentence into a single vector, which lies in a shared space between source and target languages.
Scoring a source-target sentence pair is done by computing similarity between the source embedding vector and the target embedding vector.
It is much more efficient than scoring by decoding, e.g. with a translation model.

Bilingual sentence embeddings have been studied primarily for transfer learning of monolingual downstream tasks across languages \cite{hermann2014multilingual,pham2015learning,zhou2016cross}.
However, few papers apply it to bilingual corpus mining; many of them require parallel training data with additional pivot languages \cite{espana2017empirical,schwenk2018filtering} or lack an investigation into similarity between the embeddings \cite{guo2018effective}.

This work solves these issues as follows:
\begin{itemize}\itemsep0em
    \item We propose a simple end-to-end training approach of bilingual sentence embeddings with parallel and monolingual data only of the corresponding language pair.
    \item We use a multilayer perceptron (MLP) as a trainable similarity measure to match source and target sentence embeddings.
    \item We compare various similarity measures for embeddings in terms of score distribution, geometric interpretation, and performance in downstream tasks.
    \item We demonstrate competitive performance in sentence alignment recovery and parallel corpus filtering tasks without a complex combination of translation/language models.
    \item We analyze the effect of negative examples on training an MLP similarity, using different levels of negativity.
\end{itemize}


\section{Related Work}

Bilingual representation of a sentence was at first built by averaging pre-trained bilingual word embeddings \cite{huang2012improving,klementiev2012inducing}.
The compositionality from words to sentences is integrated into end-to-end training in \newcite{hermann2014multilingual}. 

Explicit modeling of a sentence-level bilingual embedding was first discussed in \newcite{chandar2013multilingual}, training an autoencoder on monolingual sentence embeddings of two languages.
\newcite{pham2015learning} jointly learn bilingual sentence and word embeddings by feeding a shared sentence embedding to $n$-gram models.
\newcite{zhou2016cross} add document-level alignment information to this model as a constraint in training.

Recently, sequence-to-sequence NMT models were adapted to learn cross-lingual sentence embeddings.
\newcite{schwenk2017learning} connect multiple source encoders to a shared decoder of a pivot target language, forcing the consistency of encoder representations.
\newcite{schwenk2018filtering} extend this work to use a single encoder for many source languages.
Both methods rely on $N$-way parallel training data, which are seriously limited to certain languages and domains.
\newcite{artetxe2018massively} relax this data condition to pairwise parallel data including the pivot language, but it is still unrealistic for many scenarios (see Section \ref{sec:filtering}).
In contrast, our method needs only parallel and monolingual data for source and target languages of concern without any pivot languages.

\newcite{hassan2018achievingHumanPar} train a bidirectional NMT model with a single encoder-decoder, taking the average of top-layer encoder states as the sentence embedding.
They do not include any details on the data or translation performance before/after the filtering with this embedding.
\newcite{junczys2018dual} apply this method to WMT 2018 parallel corpus filtering task, yet showing significantly worse performance than a combination of translation/language models.
Our method shows comparable results to such model combinations in the same task.

\newcite{guo2018effective} replace the decoder with a feedforward network and use the parallel sentences as input to the two encoders.
Similarly to our work, the feedforward network measures the similarity of sentence pairs, except that the source and target sentence embeddings are combined via dot product instead of concatenation.
Their model, however, is not directly optimizing the source and target sentences to be translations of each other; it only attaches two encoders in the output level without a decoder.

Based on the model of \newcite{artetxe2018massively}, \newcite{artetxe2018margin} scale cosine similarity between sentence embeddings with average similarity of the nearest neighbors.
Searching for the nearest neighbors among hundreds of millions of sentences may cause a huge computational problem.
On the other hand, our similarity calculation is much quicker and support batch computation while preserving strong performance in parallel corpus filtering.

Neither of the above-mentioned methods utilize monolingual data.
We integrate autoencoding into NMT to maximize the usage of parallel and monolingual data together in learning bilingual sentence embeddings.


\section{Bilingual Sentence Embeddings}



A bilingual sentence embedding function maps sentences from both the source and target language into a single joint vector space.
Once we obtain such a space, we can search for a similar target sentence embedding given a source sentence embedding, or vice versa.


\subsection{Model}

In this work, we learn bilingual sentence embeddings via NMT and autoencoding given parallel and monolingual corpora.
Since our purpose is to pair source and target sentences, translation is a natural base task to connect sentences in two different languages.
We adopt a basic encoder-decoder approach from \newcite{sutskever2014sequence}.
The encoder produces a fixed-length embedding of a source sentence, which is used by the decoder to generate the target hypothesis.

First, the encoder takes a source sentence $f_1^J=f_1,...,f_j,...,f_J$ (length $J$) as input, where each $f_j$ is a source word.
It computes hidden representations $\mathbf{h}_j \in \mathbb{R}^{D}$ for all source positions $j$:
\begin{align}
    \mathbf{h}_1^J = \mathrm{enc}_\mathrm{src}(f_1^J)
\end{align}
$\mathrm{enc}_\mathrm{src}$ is implemented as a bidirectional recurrent neural network (RNN).
We denote a target output sentence by $e_1^I=e_1,...,e_i,...,e_I$ (length $I$).
The decoder is an unidirectional RNN whose internal state for a target position $i$ is:
\begin{align}
    \mathbf{s}_i = \mathrm{dec}(\mathbf{s}_{i-1},e_{i-1})
\end{align}
where its initial state is element-wise max-pooling of the encoder representations $\mathbf{h}_1^J$:
\begin{align}
    \mathbf{s}_0 &= \mathrm{maxpool}(\mathbf{h}_1^J)\nonumber\\
    &= \left[\max_{j=1,...,J} \mathbf{h}_{j1},\; ...\:,\; \max_{j=1,...,J} \mathbf{h}_{jD}\right]^\top
\end{align}
We empirically found that the max-pooling performs much better than averaging or choosing the first ($\mathbf{h}_1$) or last ($\mathbf{h}_J$) representation.
Finally, an output layer predicts a target word $e_i$:
\begin{align}
    p_\theta(e_i|e_1^{i-1},f_1^J) = \mathrm{softmax}(\mathrm{linear}(\mathbf{s}_i))
\end{align}
where $\theta$ denotes a set of model parameters.

Note that the decoder has access to the source sentence only through $\mathbf{s}_0$, which we take as the sentence embedding of $f_1^J$.
This assumes that the source sentence embedding contains sufficient information for translating to a target sentence, which is desired for a bilingual embedding space.

\begin{figure}[!t]
    \centering
    \begin{overpic}[width=0.45\textwidth]{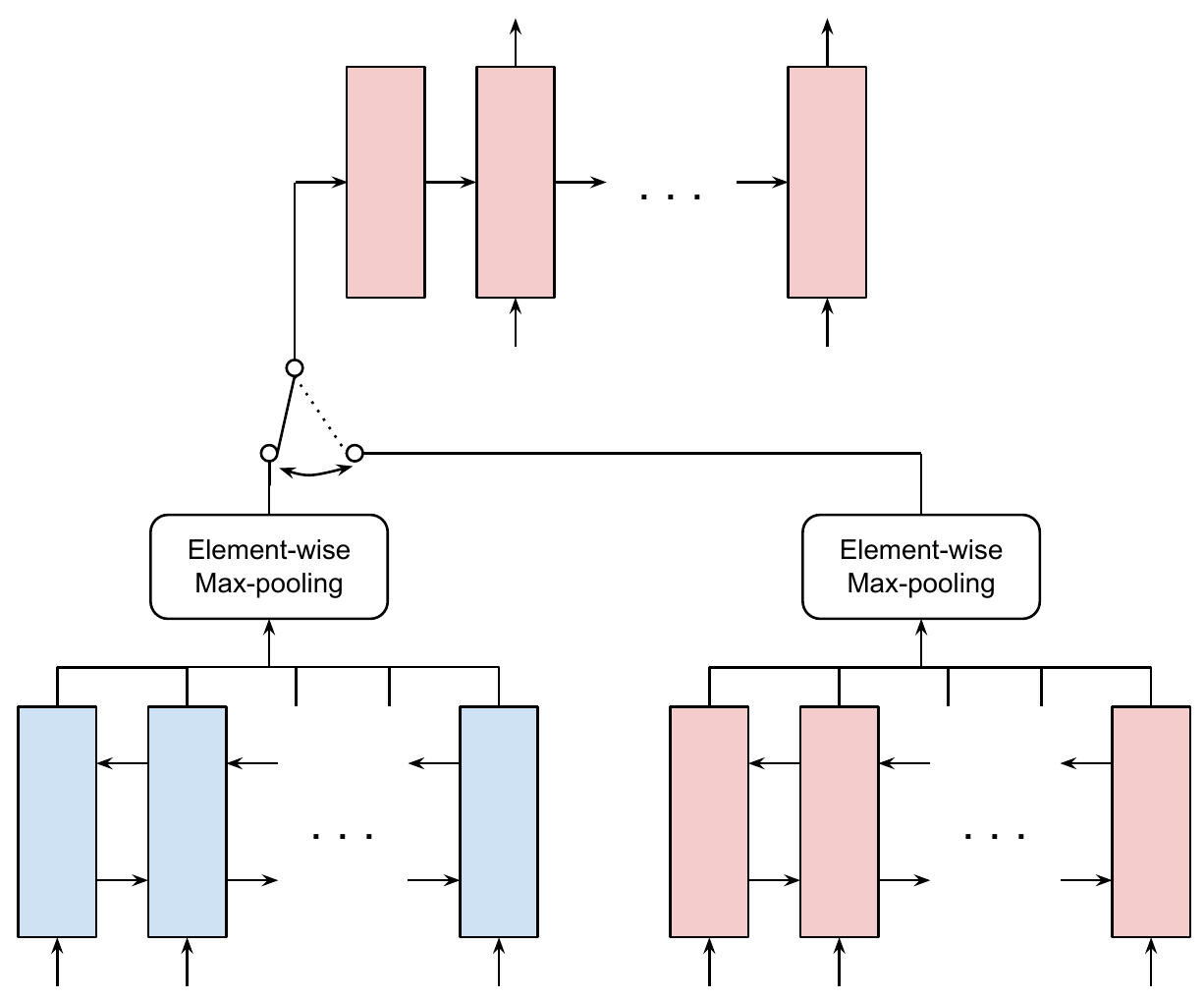}
        \put (2,-4) {$f_1$}
        \put (13,-4) {$f_2$}
        \put (39,-4) {$f_J$}
        \put (57,-4) {$e_1$}
        \put (68,-4) {$e_2$}
        \put (94,-4) {$e_I$}
        \put (2,14) {$\mathbf{h}_1$}
        \put (13,14) {$\mathbf{h}_2$}
        \put (38.5,14) {$\mathbf{h}_J$}
        \put (56.5,14) {$\bar{\mathbf{h}}_1$}
        \put (67,14) {$\bar{\mathbf{h}}_2$}
        \put (93,14) {$\bar{\mathbf{h}}_I$}
        \put (40,50) {$e_0$}
        \put (65,50) {$e_{i-1}$}
        \put (29.5,66.5) {$\mathbf{s}_0$}
        \put (40.5,66.5) {$\mathbf{s}_1$}
        \put (66.5,66.5) {$\mathbf{s}_I$}
        \put (40,83) {$e_1$}
        \put (66,83) {$e_I$}
    \end{overpic}\vspace{1em}
    \caption{Our proposed model for learning bilingual sentence embeddings. A decoder (above) is shared over two encoders (below). The decoder accepts a max-pooled representation from either one of the encoders as its first state $\mathbf{s}_0$, depending on the training objective (Equation \ref{eq:s2t-nmt} and \ref{eq:s2t-ae}).}
    \label{fig:model}\vspace{-0.5em}
\end{figure}
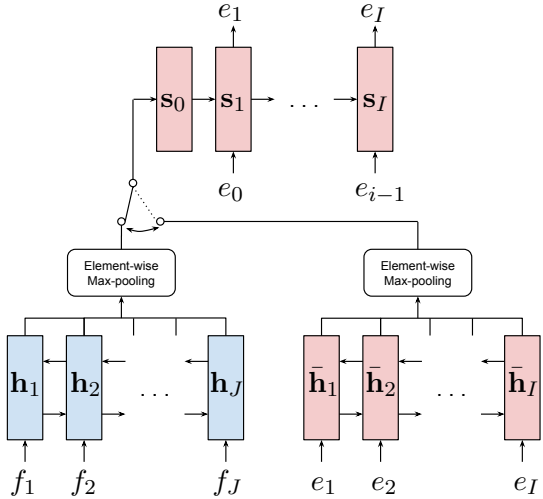

However, this plain NMT model can generate only source sentence embeddings through the encoder.
The decoder cannot process a new target sentence without a proper source language input.
We can perform decoding with an empty source input and take the last decoder state $\mathbf{s}_I$ as the sentence embedding of $e_1^I$, but it is not compatible with the source embedding and contradicts the way in which the model is trained.

Therefore, we attach another encoder of the target language to the same (target) decoder:
\begin{align}
    \bar{\mathbf{h}}_1^I &= \mathrm{enc}_\mathrm{tgt}(e_1^I)\\
    \mathbf{s}_0 &= \left[\max_{i=1,...,I} \bar{\mathbf{h}}_{i1},\; ...\:,\; \max_{i=1,...,I} \bar{\mathbf{h}}_{iD}\right]^\top
\end{align}
$\mathrm{enc}_\mathrm{tgt}$ has the same architecture as $\mathrm{enc}_\mathrm{src}$.
The model has now an additional information flow from a target input sentence to the same target (output) sentence, also known as sequential autoencoder \cite{li2015hierarchical}.

Figure \ref{fig:model} is a diagram of our model.
A decoder is shared between NMT and autoencoding parts; it takes either source or target sentence embedding and does not differentiate between the two when producing an output.
The two encoders are constrained to provide mathematically consistent representations over the languages (to the decoder).

Note that our model does not have any attention component \cite{bahdanau2014neural}.
The attention mechanism in NMT makes the decoder attend to encoder representations at all source positions.
This is counterintuitive for our purpose; we need to optimize the encoder to produce a single representation vector, but the attention model allows the encoder to distribute information over many different positions. 
In our initial experiments, the same model with the attention mechanism showed exorbitantly bad performance, so we removed it in the main experiments of Section \ref{sec:eval}.

\subsection{Training and Inference}

Let $\theta_{\mathrm{enc}_\mathrm{src}}$, $\theta_{\mathrm{enc}_\mathrm{tgt}}$, and $\theta_{\mathrm{dec}}$ the parameters of the source encoder, the target encoder, and the (shared) decoder, respectively.
Given a parallel corpus $\mathcal{P}$ and a target monolingual corpus $\mathcal{M}_\mathrm{tgt}$, the training criterion of our model is the cross-entropy on two input-output paths.
The NMT objective (Equation \ref{eq:s2t-nmt}) is for training $\theta_1 = \{\theta_{\mathrm{enc}_\mathrm{src}}, \theta_{\mathrm{dec}}\}$, and the autoencoding objective (Equation \ref{eq:s2t-ae}) is for training $\theta_2 = \{\theta_{\mathrm{enc}_\mathrm{tgt}}, \theta_{\mathrm{dec}}\}$:
\begin{align}
    L_\mathrm{emb}(\theta) = \,&-\hspace{-10pt}\sum_{(f_1^J,e_1^I)\in\mathcal{P}}\hspace{-10pt} \log p_{\theta_1}(e_1^I|f_1^J)\label{eq:s2t-nmt}\\
    &-\hspace{-8pt}\sum_{e_1^I\in\mathcal{M}_\mathrm{tgt}}\hspace{-8pt} \log p_{\theta_2}(e_1^I|e_1^I)\label{eq:s2t-ae}
\end{align}

\noindent where $\theta=\{\theta_1,\theta_2\}$. During the training, each mini-batch contains examples of the both objectives with a 1:1 ratio.
In this way, we prevent one encoder from being optimized more than the other, forcing the two encoders produce balanced sentence embeddings that fit to the same decoder.

The autoencoding part can be trained with a separate target monolingual corpus.
To provide a stronger training signal for the shared embedding space, we use also the target side of $\mathcal{P}$; the model learns to produce the same target sentence from the corresponding source and target inputs.

In order to guide the training to bilingual representations, we initialize the word embedding layers with a pre-trained bilingual word embedding.
The word embedding for each language is trained with a skip-gram algorithm \cite{mikolov2013efficient}, later mapped across the languages with adversarial training \cite{conneau2018word} and self-dictionary refinements \cite{artetxe2017learning}.

Our model can be built also in the opposite direction, i.e. with a target-to-source NMT model and a source autoencoder:
\begin{align}
    L_\mathrm{emb}(\theta) = \,&-\hspace{-10pt}\sum_{(f_1^J,e_1^I)\in\mathcal{P}}\hspace{-10pt} \log p_{\theta_2}(f_1^J|e_1^I)\label{eq:t2s-nmt}\\
    &-\hspace{-8pt}\sum_{f_1^J\in\mathcal{M}_\mathrm{src}}\hspace{-8pt} \log p_{\theta_1}(f_1^J|f_1^J)\label{eq:t2s-ae}
\end{align}

Once the model is trained, we need only the encoders to query sentence embeddings.
Let $\mathbf{a}$ and $\mathbf{b}$ be embeddings of a source sentence $f_1^J$ and a target sentence $e_1^I$, respectively:
\begin{align}
    \mathbf{a} &= \mathrm{maxpool}(\mathrm{enc}_\mathrm{src}(f_1^J))\\
    \mathbf{b} &= \mathrm{maxpool}(\mathrm{enc}_\mathrm{tgt}(e_1^I))
\end{align}

\subsection{Computing Similarities}

The next step is to evaluate how close the two embeddings are to each other, i.e. to compute a similarity measure between them.
In this paper, we consider two types of similarity measures.\vspace{0.6em}

\noindent\textbf{Predefined mathematical functions}\hspace{0.3cm}
Cosine similarity is a conventional choice for measuring the similarity in vector space modeling of information retrieval or text mining \cite{singhal2001modern}.
It computes the angle between two vectors (rotation) and ignore the lengths:
\begin{align}
    \mathrm{cos}(\mathbf{a},\mathbf{b})=\frac{\mathbf{a} \cdot \mathbf{b}}{\|\mathbf{a}\|\|\mathbf{b}\|}
\end{align}

Euclidean distance indicates how much distance must be traveled to move from the end of a vector to that of the other (transition). We reverse this distance to use it as a similarity measure:
\begin{align}
    \mathrm{Euclidean}(\mathbf{a},\mathbf{b}) = -\|\mathbf{a}-\mathbf{b}\|
\end{align}

However, these simple measures, i.e. a single rotation or transition, might not be sufficient to define the similarity of complex natural language sentences across different languages.
Also, the learned joint embedding space is not necessarily perfect in the sense of vector space geometry; even if we train it with a decent algorithm, the structure and quality of the embedding space are highly dependent on the amount of parallel training data and its domain.
This might hinder the simple functions from working well for our purpose.
\vspace{0.6em}

\noindent\textbf{Trainable multilayer perceptron}\hspace{0.3cm}
To model relations of sentence embeddings by combining rotation, shift, and even nonlinear transformations, We train a small multilayer perceptron (MLP) \cite{bishop1995neural} and use it as a similarity measure.
We design the MLP network $q(\mathbf{a},\mathbf{b})$ as a simple binary classifier whose input is a concatenation of source and target sentence embeddings: $[\mathbf{a};\mathbf{b}]^\top$.
It is passed through feedforward hidden layers with nonlinear activations.
The output layer has a single node with sigmoid activation, representing how probable the source and target sentences are translations of each other.

To train this model, we must have positive examples (real parallel sentence pairs, $\mathcal{P}_\mathrm{pos}$) and negative examples (nonparallel or noisy sentence pairs, $\mathcal{P}_\mathrm{neg}$).
The training criterion is:
\begin{align}
    L_\mathrm{sim} = \,&-\hspace{-8pt}\sum_{(\mathbf{a},\mathbf{b})\in\mathcal{P}_\mathrm{pos}}\hspace{-8pt}\log q(\mathbf{a},\mathbf{b})\nonumber\\
    &-\hspace{-8pt}\sum_{(\mathbf{a},\mathbf{b})\in\mathcal{P}_\mathrm{neg}}\hspace{-8pt}(1 - \log q(\mathbf{a},\mathbf{b}))\label{eq:loss-sim}
\end{align}
which naturally fits to the main task of interest: parallel corpus filtering (Section \ref{sec:filtering}).
Note that the output of the MLP can be quite biased to the extremes (0 or 1) in order to clearly distinguish good and bad examples.
This has both advantages and disadvantages as explained in Section \ref{analysis:measures}.

Our MLP similarity can be optimized differently for each embedding space.
Furthermore, the user can inject domain-specific knowledge into the MLP similarity by training only with in-domain parallel data.
The resulting MLP would devalue not only nonparallel sentence pairs but also out-of-domain instances.

\section{Evaluation}
\label{sec:eval}

We evaluated our bilingual sentence embedding and the MLP similarity on two tasks: sentence alignment recovery and parallel corpus filtering.
The sentence embedding was trained with WMT 2018 English-German parallel data and 100M German sentences from the News Crawl monolingual data\footnote{http://www.statmt.org/wmt18/translation-task.html}, where we use German as the autoencoded language.
All sentences were lowercased and limited to the length of 60.
We learned the byte pair encoding \cite{sennrich2016neural} jointly for the two languages with 20k merge operations.
We pre-trained bilingual word embeddings on 100M sentences from the News Crawl data for each language using \textsc{fasttext} \cite{bojanowski2017enriching} and \textsc{MUSE} \cite{conneau2018word}.


Our sentence embedding model has 1-layer RNN encoder/decoder, where the word embedding and hidden layers have a size of 512. The training was done with stochastic gradient descent with initial learning rate of 1.0, batch size of 120 sentences, and maximum 800k updates. After 100k updates, we reduced the learning rate by a factor of 0.9 for every 50k updates.

Our MLP similarity model has 2 hidden layers of size 512 with ReLU \cite{nair2010rectified}, trained with \textsc{scikit-learn} \cite{scikit-learn} with maximum 1,000 updates.
For a positive training set, we used newstest2007-2015 from WMT (around 21k sentences).
Unless otherwise noted, we took a comparable size of negative examples from the worst-scored sentence pairs of ParaCrawl\footnote{https://www.paracrawl.eu/} English-German corpus.
The scoring was done with our bilingual sentence embedding and cosine similarity.

Note that the negative examples are selected via cosine similarity but the similarity values are not used in the MLP training (Equation \ref{eq:loss-sim}).
Thus it does not learn to mimic the cosine similarity function again, but has a new sorting of sentence pairs---also encoding the domain information.


\subsection{Sentence Alignment Recovery}

In this task, we corrupt the sentence alignments of a parallel test set by shuffling one side, and find the original alignments; also known as corpus reconstruction \cite{schwenk2017learning}.

Given a source sentence, we compute a similarity score with every possible target sentence in the data and take the top-scored one as the alignment.
The error rate is the number of incorrect sentence alignments divided by the total number of sentences.
We compute this also in the opposite direction and take an average of the two error rates.
It is an intrinsic evaluation for parallel corpus mining.
We choose two test sets: WMT newstest2018 (2998 lines) and IWSLT tst2015 (1080 lines).

As baselines, we used character-level Levenshtein distance and length-normalized posterior scores of German$\rightarrow$English/English$\rightarrow$German NMT models. Each NMT model is a 3-layer base Transformer \cite{vaswani2017attention} trained on the same training data as the sentence embedding.


\begin{table}[!h]
    \centering
    \setlength\tabcolsep{4pt}
    \begin{tabular}{lcc}
    \toprule
    & \multicolumn{2}{c}{Error [\%]}\\
    \cmidrule{2-3}
    Method & WMT & IWSLT\\
    \midrule
    Levenshtein distance & 37.4 & 54.6\\
    NMT de-en + en-de & \textbf{1.7} & \textbf{13.3}\\
    \midrule
    Our method (Cosine similarity) & \textbf{4.3} & \textbf{13.8}\\
    Our method (MLP similarity) & 89.9 & 72.6\\
    \bottomrule
    \end{tabular}
    \caption{Sentence alignment recovery results. Our method results use cosine similarity except the last row.}
    \label{tab:recovery}
\end{table}

Table \ref{tab:recovery} shows the results. The Levenshtein distance gives a poor performance. NMT models are better than the other methods, but takes too long to compute posteriors for all possible pairs of source and target sentences (about 12 hours for the WMT test set). This is absolutely not feasible for a real mining task with hundreds of millions of sentences.

Our bilingual sentence embeddings (with using cosine similarity) show error rates close to the NMT models, especially in the IWSLT test set. Computing similarities between embeddings is extremely fast (about 3 minutes for the WMT test set), which perfectly fits to mining scenarios.

However, the MLP similarity performs bad in aligning sentence pairs. Given a source sentence, it puts all reasonably similar target sentences to the score 1 and does not precisely distinguish between them. Detailed investigation of this behavior is in Section \ref{analysis:measures}. As we will find out, this is ironically very effective in parallel corpus filtering.

\begin{table*}[!ht]
    \centering
    \begin{tabular}{lccccc}
    \toprule
    & \multicolumn{5}{c}{\textsc{Bleu} [\%]}\\
    & \multicolumn{2}{c}{10M words} & & \multicolumn{2}{c}{100M words}\\
    \cmidrule{2-3} \cmidrule{5-6}
    Method & test2017 & test2018 & & test2017 & test2018\\
    \midrule
    Random sampling & 19.1 & 23.1 & & 23.2 & 29.3\\
    Pivot-based embedding \cite{schwenk2017learning} & 26.1 & 32.4 & & 30.0 & 37.5\\ 
    NMT + LM, 4 models \cite{rossenbach2018rwth} & 29.1 & 35.2 & & \textbf{31.3} & \textbf{38.2}\\
    \midrule
    Our method (cosine similarity) & 23.0 & 28.4 & & 27.9 & 34.4\\
    Our method (MLP similarity) & \textbf{29.2} & \textbf{35.4} & & 30.6 & 37.5\\
    \bottomrule
    \end{tabular}
    \caption{Parallel corpus filtering results (German$\rightarrow$English).}
    \label{tab:filtering}
\end{table*}

\subsection{Parallel Corpus Filtering}
\label{sec:filtering}

We also test our methods in the WMT 2018 parallel corpus filtering task \cite{koehn2018findings}.\vspace{0.7em}

\noindent\textbf{Data}\hspace{0.3cm} The task is to score each line of a very noisy, web-crawled corpus of 104M parallel lines (ParaCrawl English-German).
We pre-filtered the given raw corpus with the heuristics of \newcite{rossenbach2018rwth}.
Only the data for WMT 2018 English-German news translation task is allowed to train scoring models.
The evaluation procedure is: subsample top-scored lines which amounts to 10M/100M words, train a small NMT model with the subsampled data, and check its translation performance.
We follow the official pipeline except that we train 3-layer Transformer NMT model using Sockeye \cite{hieber2017sockeye} for evaluation.\vspace{0.7em}


\noindent\textbf{Baselines}\hspace{0.3cm} We have three comparative baselines: 1) random sampling, 2) bilingual sentence embedding learned with a third pivot target language \cite{schwenk2017learning}, 3) combination of source-to-target/target-to-source NMT and source/target LM \cite{rossenbach2018rwth}, a top-ranked system in the official evaluation.

Note that the second method violates the official data condition of the task since it requires parallel data in German-Pivot and English-Pivot. This method is not practical when learning multilingual embeddings for English and other languages, since it is hard to collect pairwise parallel data involving a non-English pivot language (except among European languages).
We trained this method with $N$-way parallel UN corpus \cite{ziemski2016united} with French as the pivot language.
The size of this model is the same as that of our autoencoding-based model except the word embedding layers.\vspace{0.7em}

The results are shown in Table \ref{tab:filtering}, where cosine similarity was used by default for sentence embedding methods except the last row.
Pivot-based sentence embedding \cite{schwenk2017learning} improves upon the random sampling, but it has an impractical data condition. The four-model combination of NMT models and LMs \cite{rossenbach2018rwth} provide 1-3\% more \textsc{Bleu} improvement.
Note that, for the third method, each model costs 1-2 weeks to train.

Our bilingual sentence embedding method greatly improves over the random sampling baseline up to 5.3\% \textsc{Bleu} in the 10M-word case and 5.1\% \textsc{Bleu} in the 100M-word case. With our MLP similarity, the improvement in \textsc{Bleu} is up to 12.3\% and 8.2\% in the 10M-word case and the 100M-word case, respectively. It outperforms the pivot-based embedding method significantly and gets close to the performance of the four-model combination. Note that we use only a single model trained with only given parallel/monolingual data for the corresponding language pair, i.e. English-German. In contrast to sentence alignment recovery experiments, the MLP similarity boosts the filtering performance by a large margin.

\section{Analysis}

In this section, we provide more in-depth analyses to compare 1) various similarity measures and 2) different choices of the negative training set for the MLP similarity model.

\subsection{Similarity Measures}\label{analysis:measures}

\begin{table}[!ht]
    \centering
    \begin{tabular}{lccc}
    \toprule
    & \multicolumn{3}{c}{Error [\%]}\\
    Similarity & de-en & en-de & Average\\
    \midrule
    Euclidean & 7.9 & 99.8 & 53.8\\
    Cosine & 4.3 & 4.2 & 4.3\\
    CSLS & 1.9 & 2.2 & 2.1\\
    MLP & 85.0 & 94.8 & 89.9\\
    \bottomrule
    \end{tabular}
    \caption{Sentence alignment recovery results with different similarity measures (newstest2018).}
    \label{tab:measures}\vspace{-0.1em}
\end{table}

In Table \ref{tab:measures}, we compare sentence alignment recovery performance with different similarity measures.

Euclidean distance shows a worse performance than cosine similarity. This means that in a sentence embedding space, we should consider rotation more than transition when comparing two vectors. Particularly, the English$\rightarrow$German direction has a peculiarly bad result with Euclidean distance. This is due to a hubness problem in a high-dimensional space, where some vectors are highly likely to be nearest neighbors of many others.

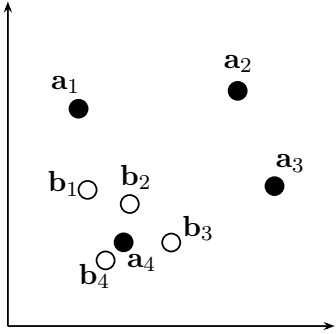
\begin{figure}[!h]
    \centering
    \begin{overpic}[width=0.28\textwidth]{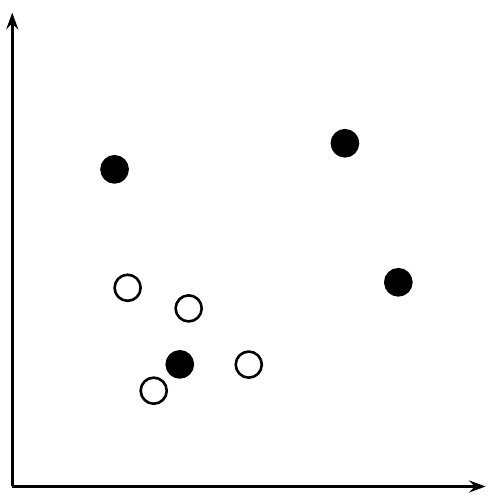}
        \put (15,72) {$\mathbf{a}_1$}
        \put (65,78) {$\mathbf{a}_2$}
        \put (80,49) {$\mathbf{a}_3$}
        \put (37,20) {$\mathbf{a}_4$}
        \put (14,42) {$\mathbf{b}_1$}
        \put (35,44) {$\mathbf{b}_2$}
        \put (53,29) {$\mathbf{b}_3$}
        \put (23,15) {$\mathbf{b}_4$}
    \end{overpic}
    \caption{Schematic diagram of the hubness problem. Filled circles indicate German sentence embeddings, while empty circles denote English sentence embeddings. All embeddings are assumed to be normalized.}
    \label{fig:hub}
\end{figure}

Figure \ref{fig:hub} illustrates that Euclidean distance is more prone to the hubs than cosine similarity. Assume that German sentence embeddings $\mathbf{a}_n$ and English sentence embeddings $\mathbf{b}_n$ should match to each other with the same index $n$, e.g. ($\mathbf{a}_1$,$\mathbf{b}_1$) is a correct match. With cosine similarity, the nearest neighbor of $\mathbf{a}_n$ is always $\mathbf{b}_n$ for all $n=1,...,4$ and vice versa, considering only the angles between the vectors. However, when using Euclidean distance, there is a discrepancy between German$\rightarrow$English and English$\rightarrow$German directions: The nearest neighbor of each $\mathbf{a}_n$ is $\mathbf{b}_n$, but the nearest neighbor of all $\mathbf{b}_n$ is always $\mathbf{a}_4$. This leads to a serious performance drop only in English$\rightarrow$German. The figure is depicted in a two-dimensional space for simplicity, but the hubness problem becomes worse for an actual high-dimensional space of sentence embeddings.

Cross-domain similarity local scaling (CSLS) is developed to counteract the hubness problem by penalizing similarity values in dense areas of the embedding distribution \cite{conneau2018word}:
\begin{align}
    \mathrm{CSLS}(\mathbf{a},\mathbf{b}) = \;&2 \cdot \mathrm{cos}(\mathbf{a},\mathbf{b})\\
    &- \frac{1}{K}\sum_{\mathbf{b}' \in \mathrm{NN}(\mathbf{a})}\hspace{-8pt}\mathrm{cos}(\mathbf{a},\mathbf{b}')\label{eq:csls-pen1}\\
    &- \frac{1}{K}\sum_{\mathbf{a}' \in \mathrm{NN}(\mathbf{b})}\hspace{-8pt}\mathrm{cos}(\mathbf{a}',\mathbf{b})\label{eq:csls-pen2}
\end{align}
where $K$ is the number of nearest neighbors. CSLS outperforms cosine similarity in our experiments. For a large-scale mining scenario, however, the measure requires heavy computations for the penalty terms (Equation \ref{eq:csls-pen1} and \ref{eq:csls-pen2}), i.e. nearest neighbor search in all combinations of source and target sentences and sorting the scores over e.g. a few hundred million instances.

\begin{figure}[!t]
\centering
\begin{subfigure}[t]{0.4\textwidth}
\centering\hspace{-25pt}
\resizebox{0.77\linewidth}{!}{
\begin{tikzpicture}
\begin{axis}[
	xlabel={sentence index ($\times 10^7$)},
	ylabel={similarity score},
	ymajorgrids=true,
	ymin=-0.01,
	ymax=1.01,
	xtick scale label code/.code={},
	legend pos=north west,
	width=\textwidth,
	height=7cm,
	]
\addplot+[mark=none, line width=0.5mm] table {similarity.cosine.sorted.dat};
\end{axis}
\end{tikzpicture}}
\caption{\label{fig:score_distribution:cosine}%
Cosine similarity}
\end{subfigure}\vspace{0.5em}
\begin{subfigure}[t]{0.4\textwidth}
\centering\hspace{-25pt}
\resizebox{0.77\textwidth}{!}{
\begin{tikzpicture}
\begin{axis}[
	xlabel={sentence index ($\times 10^7$)},
	ylabel={similarity score},
	ymajorgrids=true,
	ymin=-0.01,
	ymax=1.01,
	xtick scale label code/.code={},
	legend pos=north west,
	width=\textwidth,
	height=7cm,
	]
\addplot+[mark=none, line width=0.5mm] table {similarity.nn.sorted.dat};
\end{axis}
\end{tikzpicture}}
\caption{\label{fig:score_distribution:neural}%
MLP similarity}
\end{subfigure}
\caption{\label{fig:score_distribution}%
The score distribution of similarity measures. The sentences are sorted by their similarity scores. Cosine similarity values are linearly rescaled to $[0,1]$.}\vspace{-0.3em}
\end{figure}
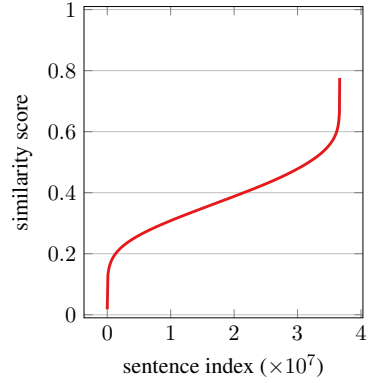
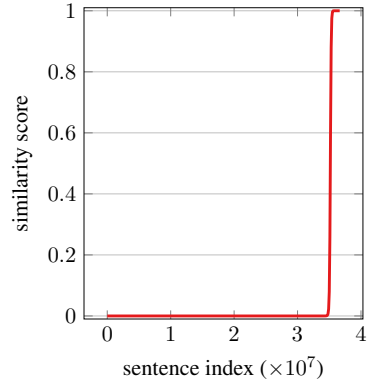

\begin{table*}[!ht]
\centering
\begin{tabular}{p{0.38\textwidth}p{0.38\textwidth}cc}
\toprule
& & \multicolumn{2}{c}{Similarity}\\
\cmidrule{3-4}
German sentence & English sentence & Cosine & MLP\\
\midrule
the requested URL / dictionary / m / \_ mar \_ eisimpleir.htm was not found on this server.&additionally, a 404 Not Found error was encountered while trying to use an ErrorDocument to handle the request. & \multirow{3}{*}{0.185}& \multirow{3}{*}{0.000} \\
\midrule
becoming Prestigious In The Right Way&how I Feel About School & 0.199 & 0.000\\
\midrule
nach dieser Aussage sollte die t\"{u}rkische Armee somit eine internationale Intervention gegen Syrien provozieren .&according to his report, the Turkish army was aiming to provoke an international intervention against Syria.&\multirow{3}{*}{0.563}&\multirow{3}{*}{1.000}\\
\midrule
allen Menschen und Besch\"{a}ftigten, die um Freiheit k\"{a}mpfen oder bei Kundgebungen ums Leben kamen, Achtung zu bezeugen und die unverz\"{u}gliche Freilassung aller Inhaftierten zu fordern&to pay tribute to all people and workers who have been fighting for freedom or fallen in demonstrations and demand the immediate release of all detainees&\multirow{5}{*}{0.427}&\multirow{5}{*}{0.999}\\
\bottomrule
\end{tabular}
\caption{\label{tab:sentex}
Example sentence pairs in the ParaCrawl corpus (Section \ref{sec:filtering}) with their similarity values.}
\end{table*}

The MLP similarity is not performing well as opposed to its results in parallel corpus filtering. To explain this, we depict score distributions of cosine and MLP similarity over the ParaCrawl corpus in Figure \ref{fig:score_distribution}. As for cosine similarity, only a small fraction of the corpus is given low- or high-range scores (smaller than 0.2 or larger than 0.6). The remaining sentences are distributed almost uniformly within the score range inbetween.

The distribution curve of the MLP similarity has a completely different shape. It has a strong tendency to classify a sentence pair to be extremely bad or extremely good: nearly 80\% of the corpus is scored with zero and only 3.25\% gets scores between 0.99 and 1.0. Table \ref{tab:sentex} shows some example sentence pairs with extreme MLP similarity values.

This is the reason why the MLP similarity does a good job in filtering, especially in selecting a small portion (10M-word) of good parallel sentences. Table \ref{tab:sentex} compares cosine similarities and the MLP scores for some sentence pairs in the raw corpus for our filtering task (Section \ref{sec:filtering}). The first two sentence pairs are absolutely nonparallel; both similarity measures give low scores, while the MLP similarity emphasizes the bad quality with zero scores. The third example is a decent parallel sentence pair with a minor ambiguity, i.e. \emph{his} in English can be a translation of \emph{dieser} in German or not, depending on the document-level context. Both measures see this sentence pair as a positive example.

The last example is parallel but the translation involves severe reordering: long-distance changes in verb positions, switching the order of relative clauses, etc. Here, cosine similarity has trouble in rating this case highly even if it is perfectly parallel, eventually filtering it out from the training data. On the other hand, our MLP similarity correctly evaluates this difficult case by giving a nearly perfect score.

However, the MLP is not optimized for precise differentiation among the good parallel matches. It is thus not appropriate for sentence alignment recovery that requires exact 1-1 matching of potential source-target pairs. A steep drop in the curve of Figure \ref{fig:score_distribution:neural} also explains why it performs slightly inferior to the best system in the 100M-word filtering task (Table \ref{tab:filtering}). The subsampling exceeds the dropping region and includes many zero-scored sentence pairs, where the MLP similarity cannot measure the quality well.

\subsection{Negative Training Examples}

In the MLP similarity training, we can use publicly available parallel corpora as the positive sets. For the negative sets, however, it is not clear which dataset we should use: entirely nonparallel sentences, partly parallel sentences, or sentence pairs of quality inbetween. We experimented with negative examples of different quality in Table \ref{tab:negativity}. 
Here is how we vary the negativity:
\begin{enumerate}\itemsep0em
    \item Score the sentence pairs of the ParaCrawl corpus with our bilingual sentence embedding using cosine similarity.
    \item Sort the sentence pairs by the scores.
    \item Divide the sorted corpus into five portions by top-scored cut of 20\%, 40\%, 60\%, 80\%, and 100\%.
    \item Take the last 100k lines for each portion. 
\end{enumerate}

A negative set from the 20\%-worst part stands for relatively less problematic sentence pairs, intending for elaborate classification among perfect parallel sentences (positive set) and almost perfect ones. With the 100\%-worst examples, we focus on removing absolutely nonsense pairing of sentences. As a simple baseline, we also take 100k sentences randomly without scoring, representing mixed levels of negativity.

\begin{table}[!t]
    \centering
    \begin{tabular}{lc}
    \toprule
    Negative examples & \textsc{Bleu} [\%]\\
    \midrule
    Random sampling & 33.3\\
    \midrule
    20\% worst & 29.9\\
    40\% worst & 33.3\\
    60\% worst & \textbf{33.7}\\
    80\% worst & 32.1\\
    100\% worst & 25.7\\
    \bottomrule
    \end{tabular}
    \caption{Parallel corpus filtering results (10M-word task) with different negative sets for training MLP similarity (newstest2016, i.e. the validation set).}
    \label{tab:negativity}
\end{table}

The results in Table \ref{tab:negativity} show that a moderate level of negativity (60\%-worst) is most suitable for training an MLP similarity model. If the negative set contains too many excellent examples, the model may mark acceptable parallel sentence pairs with zero scores. If the negative set consists only of certainly nonparallel sentence pairs, the model is weak in discriminating mid-quality instances, some of which are crucial to improve the translation system.

Random selection of sentence pairs also works surprisingly well compared to carefully tailored negative sets. It does not require us to score and sort the raw corpus, so it is very efficient, sacrificing performance slightly. We hypothesize that the average negative level of this random set is also moderate and similar to that of the 60\%-worst.

\section{Conclusion}

In this work, we present a simple method to train bilingual sentence embeddings by combining vanilla RNN NMT (without attention component) and sequential autoencoder. By optimizing a shared decoder with combined training objectives, we force the source and target sentence embeddings to share their space. Our model is trained with parallel and monolingual data of the corresponding language pair, with neither pivot languages nor $N$-way parallel data. We also propose to use a binary classification MLP as a similarity measure for matching source and target sentence embeddings.

Our bilingual sentence embeddings show consistently strong performance in both sentence alignment recovery and the WMT 2018 parallel corpus filtering tasks with only a single model. We compare various similarity measures for bilingual sentence matching, verifying that cosine similarity is preferred for a mining task and our MLP similarity is very effective in a filtering task. We also show that a moderate level of negativity is appropriate for training the MLP similarity, using either random examples or mid-range scored examples from a noisy parallel corpus.

Future work would be regularizing the MLP training to obtain a smoother distribution of the similarity scores, which could supplement the weakness of the MLP similarity (Section \ref{analysis:measures}). Furthermore, we plan to adjust our learning procedure towards the downstream tasks, e.g. with an additional training objective to maximize the cosine similarity between the source and target encoders \cite{arivazhagan2019missing}. Our method should be tested also on many other language pairs which do not have parallel data involving a pivot language.

\section*{Acknowledgments}

\begin{center}
\vspace{0.5em}
\includegraphics[align=c,width=0.2\textwidth]{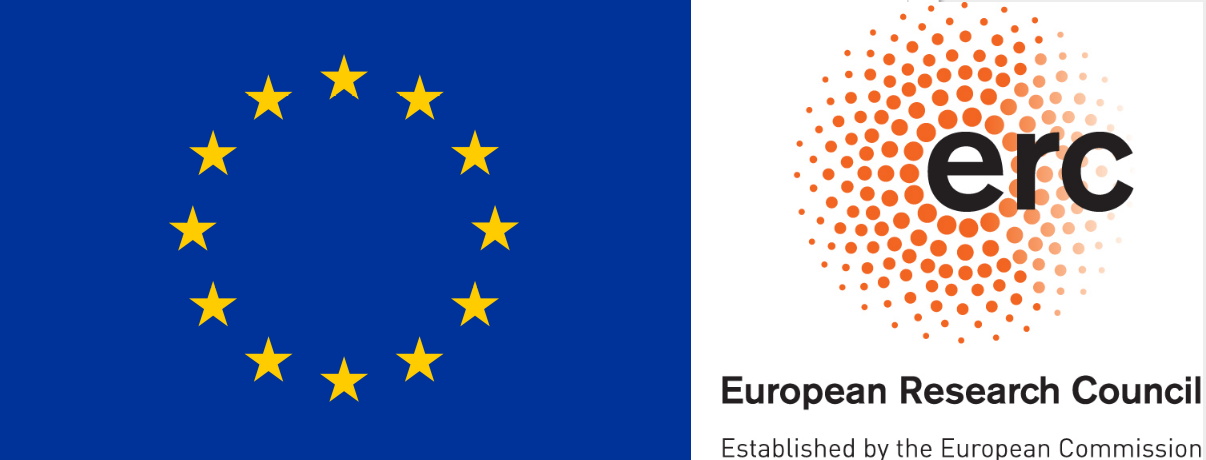}
\hspace{6pt}
\includegraphics[align=c,width=0.09\textwidth]{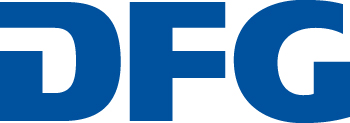}
\hspace{4pt}
\includegraphics[align=c,width=0.13\textwidth]{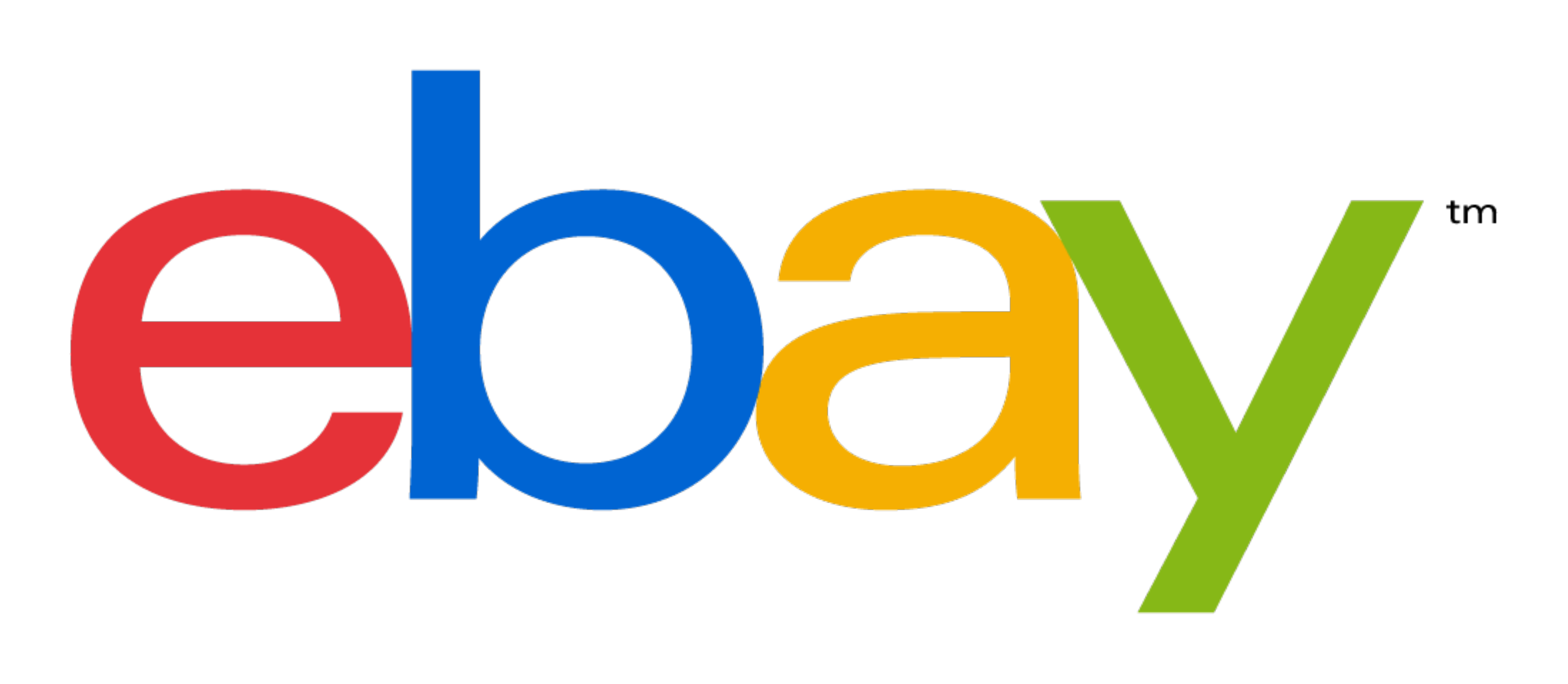}
\end{center}
\vspace{0.5em}
This work has received funding from the European Research Council (ERC) (under the European Union's Horizon 2020 research and innovation programme, grant agreement No 694537, project "SEQCLAS"), the Deutsche Forschungsgemeinschaft (DFG; grant agreement NE 572/8-1, project "CoreTec"), and eBay Inc. The GPU cluster used for the experiments was partially funded by DFG Grant INST 222/1168-1. The work reflects only the authors' views and none of the funding agencies is responsible for any use that may be made of the information it contains.

\bibliographystyle{acl_natbib}
\bibliography{references}

\begin{thebibliography}{43}
\expandafter\ifx\csname natexlab\endcsname\relax\def\natexlab#1{#1}\fi

\bibitem[{Arivazhagan et~al.(2019)Arivazhagan, Bapna, Firat, Aharoni, Johnson,
  and Macherey}]{arivazhagan2019missing}
Naveen Arivazhagan, Ankur Bapna, Orhan Firat, Roee Aharoni, Melvin Johnson, and
  Wolfgang Macherey. 2019.
\newblock The missing ingredient in zero-shot neural machine translation.
\newblock \emph{arXiv:1903.07091}.

\bibitem[{Artetxe et~al.(2017)Artetxe, Labaka, and
  Agirre}]{artetxe2017learning}
Mikel Artetxe, Gorka Labaka, and Eneko Agirre. 2017.
\newblock Learning bilingual word embeddings with (almost) no bilingual data.
\newblock In \emph{Proceedings of the 55th Annual Meeting of the Association
  for Computational Linguistics (ACL 2017)}, volume~1, pages 451--462.

\bibitem[{Artetxe and Schwenk(2018{\natexlab{a}})}]{artetxe2018margin}
Mikel Artetxe and Holger Schwenk. 2018{\natexlab{a}}.
\newblock Margin-based parallel corpus mining with multilingual sentence
  embeddings.
\newblock \emph{arXiv:1811.01136}.

\bibitem[{Artetxe and Schwenk(2018{\natexlab{b}})}]{artetxe2018massively}
Mikel Artetxe and Holger Schwenk. 2018{\natexlab{b}}.
\newblock Massively multilingual sentence embeddings for zero-shot
  cross-lingual transfer and beyond.
\newblock \emph{arXiv:1812.10464}.

\bibitem[{Bahdanau et~al.(2014)Bahdanau, Cho, and Bengio}]{bahdanau2014neural}
Dzmitry Bahdanau, Kyunghyun Cho, and Yoshua Bengio. 2014.
\newblock Neural machine translation by jointly learning to align and
  translate.
\newblock \emph{arXiv}, pages arXiv--1409.

\bibitem[{Bishop et~al.(1995)}]{bishop1995neural}
Christopher~M Bishop et~al. 1995.
\newblock \emph{Neural networks for pattern recognition}.
\newblock Oxford university press.

\bibitem[{Bojanowski et~al.(2017)Bojanowski, Grave, Joulin, and
  Mikolov}]{bojanowski2017enriching}
Piotr Bojanowski, Edouard Grave, Armand Joulin, and Tomas Mikolov. 2017.
\newblock Enriching word vectors with subword information.
\newblock \emph{Transactions of the Association for Computational Linguistics},
  5:135--146.

\bibitem[{Bojar et~al.(2018)Bojar, Federmann, Fishel, Graham, Haddow, Huck,
  Koehn, and Monz}]{bojar-EtAl:2018:WMT1}
Ond\v{r}ej Bojar, Christian Federmann, Mark Fishel, Yvette Graham, Barry
  Haddow, Matthias Huck, Philipp Koehn, and Christof Monz. 2018.
\newblock Findings of the 2018 conference on machine translation (wmt18).
\newblock In \emph{Proceedings of the Third Conference on Machine Translation,
  Volume 2: Shared Task Papers}, pages 272--307, Belgium, Brussels.

\bibitem[{Chandar et~al.(2013)Chandar, Khapra, Ravindran, Raykar, and
  Saha}]{chandar2013multilingual}
AP~Sarath Chandar, Mitesh~M Khapra, Balaraman Ravindran, Vikas Raykar, and
  Amrita Saha. 2013.
\newblock Multilingual deep learning.
\newblock In \emph{Deep Learning Workshop at NIPS}.

\bibitem[{Conneau et~al.(2018)Conneau, Lample, Ranzato, Denoyer, and
  J{\'e}gou}]{conneau2018word}
Alexis Conneau, Guillaume Lample, Marc'Aurelio Ranzato, Ludovic Denoyer, and
  Herv{\'e} J{\'e}gou. 2018.
\newblock Word translation without parallel data.
\newblock In \emph{Proceedings of 6th International Conference on Learning
  Representations (ICLR 2018)}.

\bibitem[{\d{S}tef{\u{a}}nescu et~al.(2012)\d{S}tef{\u{a}}nescu, Ion, and
  Hunsicker}]{stefuanescu2012hybrid}
Dan \d{S}tef{\u{a}}nescu, Radu Ion, and Sabine Hunsicker. 2012.
\newblock Hybrid parallel sentence mining from comparable corpora.
\newblock In \emph{Proceedings of the 16th Conference of the European
  Association for Machine Translation}, pages 137--144.

\bibitem[{Espana-Bonet et~al.(2017)Espana-Bonet, Varga, Barr{\'o}n-Cede{\~n}o,
  and van Genabith}]{espana2017empirical}
Cristina Espana-Bonet, Ad{\'a}m~Csaba Varga, Alberto Barr{\'o}n-Cede{\~n}o, and
  Josef van Genabith. 2017.
\newblock An empirical analysis of nmt-derived interlingual embeddings and
  their use in parallel sentence identification.
\newblock \emph{IEEE Journal of Selected Topics in Signal Processing},
  11(8):1340--1350.

\bibitem[{Espl\'{a}-Gomis and Forcada(2009)}]{esplagomis2009bitextor}
Miquel Espl\'{a}-Gomis and Mikel~L. Forcada. 2009.
\newblock Bitextor, a free/open-source software to harvest translation memories
  from multilingual websites.
\newblock In \emph{Proceedings of MT Summit XII}, Ottawa, Canada.

\bibitem[{Etchegoyhen and Azpeitia(2016)}]{etchegoyhen2016set}
Thierry Etchegoyhen and Andoni Azpeitia. 2016.
\newblock Set-theoretic alignment for comparable corpora.
\newblock In \emph{Proceedings of the 54th Annual Meeting of the Association
  for Computational Linguistics (Volume 1: Long Papers)}, volume~1, pages
  2009--2018.

\bibitem[{Guo et~al.(2018)Guo, Shen, Yang, Ge, Cer, Abrego, Stevens, Constant,
  Sung, Strope et~al.}]{guo2018effective}
Mandy Guo, Qinlan Shen, Yinfei Yang, Heming Ge, Daniel Cer, Gustavo~Hernandez
  Abrego, Keith Stevens, Noah Constant, Yun-hsuan Sung, Brian Strope, et~al.
  2018.
\newblock Effective parallel corpus mining using bilingual sentence embeddings.
\newblock In \emph{Proceedings of the Third Conference on Machine Translation:
  Research Papers}, pages 165--176.

\bibitem[{Hassan et~al.(2018)Hassan, Aue, Chen, Chowdhary, Clark, Federmann,
  Huang, Junczys-Dowmunt, Lewis, Li et~al.}]{hassan2018achievingHumanPar}
Hany Hassan, Anthony Aue, Chang Chen, Vishal Chowdhary, Jonathan Clark,
  Christian Federmann, Xuedong Huang, Marcin Junczys-Dowmunt, William Lewis,
  Mu~Li, et~al. 2018.
\newblock Achieving human parity on automatic chinese to english news
  translation.
\newblock \emph{arXiv preprint arXiv:1803.05567}.

\bibitem[{Hermann and Blunsom(2014)}]{hermann2014multilingual}
Karl~Moritz Hermann and Phil Blunsom. 2014.
\newblock Multilingual models for compositional distributed semantics.
\newblock In \emph{Proceedings of the 52nd Annual Meeting of the Association
  for Computational Linguistics (Volume 1: Long Papers)}, volume~1, pages
  58--68.

\bibitem[{Hieber et~al.(2017)Hieber, Domhan, Denkowski, Vilar, Sokolov,
  Clifton, and Post}]{hieber2017sockeye}
Felix Hieber, Tobias Domhan, Michael Denkowski, David Vilar, Artem Sokolov, Ann
  Clifton, and Matt Post. 2017.
\newblock Sockeye: A toolkit for neural machine translation.
\newblock \emph{arXiv preprint arXiv:1712.05690}.

\bibitem[{Huang et~al.(2012)Huang, Socher, Manning, and
  Ng}]{huang2012improving}
Eric~H Huang, Richard Socher, Christopher~D Manning, and Andrew~Y Ng. 2012.
\newblock Improving word representations via global context and multiple word
  prototypes.
\newblock In \emph{Proceedings of the 50th Annual Meeting of the Association
  for Computational Linguistics: Long Papers-Volume 1}, pages 873--882.
  Association for Computational Linguistics.

\bibitem[{Junczys-Dowmunt(2018)}]{junczys2018dual}
Marcin Junczys-Dowmunt. 2018.
\newblock Dual conditional cross-entropy filtering of noisy parallel corpora.
\newblock In \emph{Proceedings of the Third Conference on Machine Translation:
  Shared Task Papers}, pages 888--895.

\bibitem[{Kaufmann(2012)}]{kaufmann2012jmaxalign}
Max Kaufmann. 2012.
\newblock Jmaxalign: A maximum entropy parallel sentence alignment tool.
\newblock \emph{Proceedings of COLING 2012: Demonstration Papers}, pages
  277--288.

\bibitem[{Klementiev et~al.(2012)Klementiev, Titov, and
  Bhattarai}]{klementiev2012inducing}
Alexandre Klementiev, Ivan Titov, and Binod Bhattarai. 2012.
\newblock Inducing crosslingual distributed representations of words.
\newblock In \emph{Proceedings of COLING 2012}, pages 1459--1474.

\bibitem[{Koehn et~al.(2018)Koehn, Khayrallah, Heafield, and
  Forcada}]{koehn2018findings}
Philipp Koehn, Huda Khayrallah, Kenneth Heafield, and Mikel~L Forcada. 2018.
\newblock Findings of the wmt 2018 shared task on parallel corpus filtering.
\newblock In \emph{Proceedings of the Third Conference on Machine Translation:
  Shared Task Papers}, pages 726--739.

\bibitem[{Koehn and Knowles(2017)}]{koehn2017six}
Philipp Koehn and Rebecca Knowles. 2017.
\newblock Six challenges for neural machine translation.
\newblock In \emph{Proceedings of the 1st ACL Workshop on Neural Machine
  Translation (WNMT 2017)}, pages 28--39.

\bibitem[{Li et~al.(2015)Li, Luong, and Jurafsky}]{li2015hierarchical}
Jiwei Li, Thang Luong, and Dan Jurafsky. 2015.
\newblock A hierarchical neural autoencoder for paragraphs and documents.
\newblock In \emph{Proceedings of the 53rd Annual Meeting of the Association
  for Computational Linguistics and the 7th International Joint Conference on
  Natural Language Processing (Volume 1: Long Papers)}, volume~1, pages
  1106--1115.

\bibitem[{Mikolov et~al.(2013)Mikolov, Chen, Corrado, and
  Dean}]{mikolov2013efficient}
Tomas Mikolov, Kai Chen, Greg Corrado, and Jeffrey Dean. 2013.
\newblock Efficient estimation of word representations in vector space.
\newblock \emph{arXiv preprint arXiv:1301.3781}.

\bibitem[{Moore(2002)}]{moore2002fast}
Robert~C Moore. 2002.
\newblock Fast and accurate sentence alignment of bilingual corpora.
\newblock In \emph{Conference of the Association for Machine Translation in the
  Americas}, pages 135--144. Springer.

\bibitem[{Nair and Hinton(2010)}]{nair2010rectified}
Vinod Nair and Geoffrey~E Hinton. 2010.
\newblock Rectified linear units improve restricted boltzmann machines.
\newblock In \emph{Proceedings of the 27th international conference on machine
  learning (ICML-10)}, pages 807--814.

\bibitem[{Pedregosa et~al.(2011)Pedregosa, Varoquaux, Gramfort, Michel,
  Thirion, Grisel, Blondel, Prettenhofer, Weiss, Dubourg, Vanderplas, Passos,
  Cournapeau, Brucher, Perrot, and Duchesnay}]{scikit-learn}
F.~Pedregosa, G.~Varoquaux, A.~Gramfort, V.~Michel, B.~Thirion, O.~Grisel,
  M.~Blondel, P.~Prettenhofer, R.~Weiss, V.~Dubourg, J.~Vanderplas, A.~Passos,
  D.~Cournapeau, M.~Brucher, M.~Perrot, and E.~Duchesnay. 2011.
\newblock Scikit-learn: Machine learning in {P}ython.
\newblock \emph{Journal of Machine Learning Research}, 12:2825--2830.

\bibitem[{Pham et~al.(2015)Pham, Luong, and Manning}]{pham2015learning}
Hieu Pham, Thang Luong, and Christopher Manning. 2015.
\newblock Learning distributed representations for multilingual text sequences.
\newblock In \emph{Proceedings of the 1st Workshop on Vector Space Modeling for
  Natural Language Processing}, pages 88--94.

\bibitem[{Resnik and Smith(2003)}]{resnik2003web}
Philip Resnik and Noah~A Smith. 2003.
\newblock The web as a parallel corpus.
\newblock \emph{Computational Linguistics}, 29(3).

\bibitem[{Rossenbach et~al.(2018)Rossenbach, Rosendahl, Kim, Gra{\c{c}}a,
  Gokrani, and Ney}]{rossenbach2018rwth}
Nick Rossenbach, Jan Rosendahl, Yunsu Kim, Miguel Gra{\c{c}}a, Aman Gokrani,
  and Hermann Ney. 2018.
\newblock The rwth aachen university filtering system for the wmt 2018 parallel
  corpus filtering task.
\newblock In \emph{Proceedings of the Third Conference on Machine Translation:
  Shared Task Papers}, pages 946--954.

\bibitem[{Schwenk(2018)}]{schwenk2018filtering}
Holger Schwenk. 2018.
\newblock Filtering and mining parallel data in a joint multilingual space.
\newblock In \emph{Proceedings of the 56th Annual Meeting of the Association
  for Computational Linguistics (Volume 2: Short Papers)}, pages 228--234.

\bibitem[{Schwenk and Douze(2017)}]{schwenk2017learning}
Holger Schwenk and Matthijs Douze. 2017.
\newblock Learning joint multilingual sentence representations with neural
  machine translation.
\newblock In \emph{Proceedings of the 2nd Workshop on Representation Learning
  for NLP}, pages 157--167.

\bibitem[{Sennrich et~al.(2016)Sennrich, Haddow, and
  Birch}]{sennrich2016neural}
Rico Sennrich, Barry Haddow, and Alexandra Birch. 2016.
\newblock Neural machine translation of rare words with subword units.
\newblock In \emph{Proceedings of the 54th Annual Meeting of the Association
  for Computational Linguistics (Volume 1: Long Papers)}, volume~1, pages
  1715--1725.

\bibitem[{Singhal(2001)}]{singhal2001modern}
Amit Singhal. 2001.
\newblock Modern information retrieval: A brief overview.
\newblock \emph{Bulletin of the Technical Committee on}, page~35.

\bibitem[{Sutskever et~al.(2014)Sutskever, Vinyals, and
  Le}]{sutskever2014sequence}
Ilya Sutskever, Oriol Vinyals, and Quoc~V Le. 2014.
\newblock Sequence to sequence learning with neural networks.
\newblock In \emph{Proceedings of the 27th International Conference on Neural
  Information Processing Systems-Volume 2}, pages 3104--3112. MIT Press.

\bibitem[{Taghipour et~al.(2011)Taghipour, Khadivi, and
  Xu}]{taghipour2011parallel}
Kaveh Taghipour, Shahram Khadivi, and Jia Xu. 2011.
\newblock Parallel corpus refinement as an outlier detection algorithm.
\newblock In \emph{Proceedings of the 13th Machine Translation Summit (MT
  Summit XIII)}, pages 414--421.

\bibitem[{Varga et~al.(2005)Varga, Kornai, Nagy, N{\'e}meth, and
  Tr{\'o}n}]{varga2005parallel}
D{\'a}niel Varga, Andr{\'a}s Kornai, Viktor Nagy, L{\'a}szl{\'o} N{\'e}meth,
  and Viktor Tr{\'o}n. 2005.
\newblock Parallel corpora for medium density languages.
\newblock In \emph{Proceedings of RANLP 2005}, pages 590--596.

\bibitem[{Vaswani et~al.(2017)Vaswani, Shazeer, Parmar, Uszkoreit, Jones,
  Gomez, Kaiser, and Polosukhin}]{vaswani2017attention}
Ashish Vaswani, Noam Shazeer, Niki Parmar, Jakob Uszkoreit, Llion Jones,
  Aidan~N Gomez, {\L}ukasz Kaiser, and Illia Polosukhin. 2017.
\newblock Attention is all you need.
\newblock In \emph{Advances in Neural Information Processing Systems}, pages
  5998--6008.

\bibitem[{Xu and Koehn(2017)}]{xu2017zipporah}
Hainan Xu and Philipp Koehn. 2017.
\newblock Zipporah: a fast and scalable data cleaning system for noisy
  web-crawled parallel corpora.
\newblock In \emph{Proceedings of the 2017 Conference on Empirical Methods in
  Natural Language Processing}, pages 2945--2950.

\bibitem[{Zhou et~al.(2016)Zhou, Wan, and Xiao}]{zhou2016cross}
Xinjie Zhou, Xiaojun Wan, and Jianguo Xiao. 2016.
\newblock Cross-lingual sentiment classification with bilingual document
  representation learning.
\newblock In \emph{Proceedings of the 54th Annual Meeting of the Association
  for Computational Linguistics (Volume 1: Long Papers)}, volume~1, pages
  1403--1412.

\bibitem[{Ziemski et~al.(2016)Ziemski, Junczys-Dowmunt, and
  Pouliquen}]{ziemski2016united}
Michal Ziemski, Marcin Junczys-Dowmunt, and Bruno Pouliquen. 2016.
\newblock The united nations parallel corpus v1.0.
\newblock In \emph{Proceedings of Language Resources and Evaluation (LREC
  2016)}, Portoro\u{z}, Slovenia.

\end{thebibliography}

\end{document}